\documentclass[conference]{IEEEtran}
\IEEEoverridecommandlockouts

\usepackage{cite}
\usepackage{amsmath,amssymb,amsfonts}
\usepackage{algorithmic}
\usepackage{algorithm}
\usepackage{graphicx}
\usepackage{textcomp}
\usepackage{xcolor}
\usepackage{booktabs}
\usepackage{array}
\usepackage{multirow}
\usepackage{url}
\usepackage[hidelinks]{hyperref}

\begin{document}

\title{Agentic-DuplexGen: Decoupling Content, Timing, and Acoustics for Synthetic Dialogue Speech}

\author{
\IEEEauthorblockN{
Pengcheng Wang\IEEEauthorrefmark{1},
Sheng Li\IEEEauthorrefmark{1},
Jiyi Li\IEEEauthorrefmark{2},
and Takahiro Shinozaki\IEEEauthorrefmark{1}
}

\IEEEauthorblockA{
\IEEEauthorrefmark{1}
Department of Information and Communications Engineering,\\
Institute of Science Tokyo, Yokohama, Japan
}

\IEEEauthorblockA{
\IEEEauthorrefmark{2}
Hokkaido University, Hokkaido, Japan
}
}

\maketitle

\begin{abstract}
Synthetic conversational speech has become an important resource for developing and evaluating conversational speech systems. However, existing dialogue synthesis pipelines typically generate dialogue content first and then insert interruptions, overlap, and backchannels using handcrafted markers or timing rules, making conversational timing prescribed rather than interaction-driven. We present Agentic-DuplexGen, a dialogue synthesis framework that explicitly decouples \emph{content}, \emph{timing}, and \emph{acoustics}. An LLM first generates the dialogue script, and then two full-duplex conversational models perform the script while listening to each other in real time. This allows conversational timing to emerge naturally while preserving the scripted content. Finally, a high-fidelity text-to-speech model re-renders the interaction without altering its timing. As a demonstration of the proposed framework, we construct a patient--clinician conversational speech corpus with construction-time annotations, including word timestamps, speaker activity, overlap regions, and interaction events. Experimental results show that the proposed framework produces conversational dynamics closer to the real-dialogue reference distribution than the stitching baselines evaluated in this study.
\end{abstract}


% ==========================================================================
\section{Introduction}

Synthetic conversational speech has become an increasingly important resource for developing conversational speech recognition and speaker diarization systems. Recent progress in large language models (LLMs) and neural text-to-speech (TTS) has made it possible to generate large-scale dialogue corpora automatically, significantly reducing the cost of collecting human conversations. However, many existing pipelines still treat dialogue generation as a single process: the dialogue and timing are generated together, while overlap, interruption, and backchannel behaviors are subsequently introduced through markers or handcrafted timing rules~\cite{behavior_sd,personaplex}. As a result, conversational dynamics are largely prescribed by the generation pipeline instead of emerging from interaction.

Recent full-duplex conversational models naturally generate conversational interaction through continuous listening while speaking. However, these models generate conversations freely and cannot reliably follow a predefined dialogue script. Existing approaches often emphasize either controllable content or emergent interaction, leaving their combination relatively underexplored.

Our hypothesis is that dialogue generation should be decomposed into three independent processes: semantic generation, interaction generation, and acoustic realization. We address this limitation by explicitly decoupling dialogue generation into three stages: \emph{content}, \emph{timing}, and \emph{acoustics}. An LLM determines only the dialogue content and speaker order. Two full-duplex conversational models then perform the script while listening to each other in real time, allowing conversational dynamics to emerge naturally without altering the scripted dialogue. Finally, a Conditional TTS system re-renders the generated interaction while preserving the timing. This separation ensures that each stage is responsible for exactly one aspect of the conversation. 

We demonstrate this framework as \textbf{MedDialSpeech} for patient--clinician dialogue generation. Compared with the evaluated stitching baselines, the proposed approach produces conversational dynamics closer to real clinical conversations while maintaining exact script fidelity and high-quality speech synthesis. Besides, the generation process naturally produces aligned transcripts, speaker activity and interaction events, which makes the resulting corpus directly applicable to conversational ASR and speaker diarization research.

The main contributions of this work are as follows:

\begin{itemize}
\item We propose a dialogue synthesis framework that explicitly decouples dialogue content, conversational dynamics, and acoustic rendering.

\item We introduce a script-constrained duplex decoding strategy that enforces exact lexical constraints during duplex generation while allowing overlap, interruptions, and backchannels to emerge naturally through real-time interaction.

\item We construct an interaction-aware patient-clinician conversational speech corpus with construction-time annotations, and demonstrate that the proposed framework closer to the reference distribution than the evaluated stitching baselines.
\end{itemize}

% ==========================================================================
\section{Method}
\begin{figure*}[t]
\centering
\includegraphics[width=\textwidth]{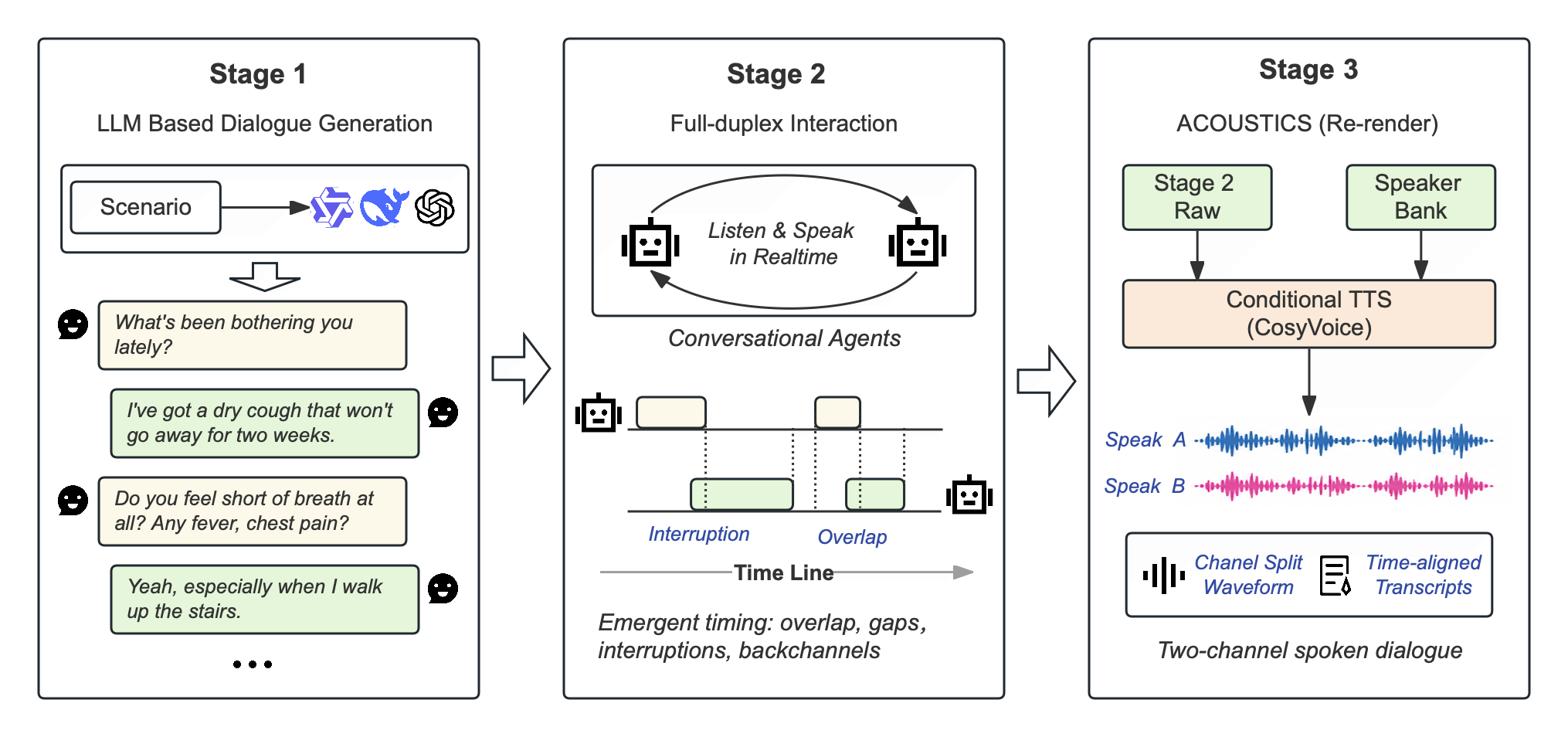}
\caption{Overview of the proposed dialogue synthesis framework. An LLM first generates the dialogue script (content). Two full-duplex conversational models then perform the script through real-time interaction, allowing conversational dynamics to emerge naturally. Finally, a high-fidelity TTS system re-renders the interaction while preserving the generated dynamics.}
\label{fig:pipeline}
\end{figure*}

% ==========================================================================
Fig.~\ref{fig:pipeline} shows the full pipeline. Its organizing principle is that content, timing, and acoustics are produced by three separate stages that never share a decision: the script fixes content, the duplex performance fixes timing, and re-rendering fixes acoustics.

\subsection{Overview and Notation}

The proposed framework separates dialogue generation into three independent stages, each responsible for one aspect of the conversation. An LLM generates the dialogue script (\emph{content}), two full-duplex conversational models perform the script through real-time interaction (\emph{timing}), and a high-fidelity TTS system re-renders the interaction (\emph{acoustics}). The three stages are strictly decoupled: the script determines only what is spoken, duplex interaction determines only when it is spoken, and re-rendering determines only how it sounds.

Let the dialogue script consist of an ordered sequence of turns assigned to two speakers, $i\in\{A,B\}$. Concatenating all turns belonging to speaker $i$ gives a target token sequence

\[
s^{i}=(s^{i}_{1},\ldots,s^{i}_{K_i}),
\]

with a cursor $p_i$ indicating the next token to be spoken. During duplex generation, both speakers continuously receive each other's audio and independently decide whether to emit the next script token or remain silent. The transcript is therefore fixed by the script, whereas the sequence of emit-or-wait decisions determines the conversational timing.

\subsection{Script-Constrained Duplex Generation}
\label{sec:m1}

Dialogue content is fixed by constraining the lexical output of the full-duplex model. Following the inner-monologue architecture of Moshi~\cite{moshi}, the model predicts a text token before generating the corresponding audio frame. At each frame, we intercept this prediction and restrict the candidate vocabulary to

\begin{equation}
\mathcal{C}^{i}_{t}=
\begin{cases}
\{\texttt{PAD}\}\cup\{s^{i}_{p_i}\},& i \text{ is floor-eligible},\\[2pt]\{\texttt{PAD}\}\cup\mathcal{B},
& i \text{ is a listener},
\end{cases}
\end{equation}

where $s^{i}_{p_i}$ is the next script token and $\mathcal{B}$ is a fixed whitelist of listener backchannels (e.g., \emph{mm-hmm}, \emph{yeah}, \emph{right}).

At each frame, the model selects

\[
w^{i}_{t}=\arg\max_{w\in\mathcal{C}^{i}_{t}}P(w\mid h^{i}_{t}).
\]

If $w^{i}_{t}=s^{i}_{p_i}$, the corresponding audio is emitted and the cursor is advanced,

\[
p_i \leftarrow p_i+1.
\]

If $w^{i}_{t}=\texttt{PAD}$, the speaker remains silent for the current frame. If $w^{i}_{t}\in\mathcal{B}$, a listener backchannel is produced subject to the refractory constraint introduced in Section~\ref{sec:m2}.

The constraint removes lexical uncertainty from duplex generation. Whenever a speaker decides to speak, the emitted word is uniquely determined by the script. Consequently, the duplex model no longer decides \emph{what} to say; it decides only \emph{whether} to emit the next script token at the current frame or to wait. Dialogue content is therefore fixed by construction, while conversational timing remains completely determined by the interaction between the two duplex instances.

\subsection{Emergent Timing}
\label{sec:m2}

Conversational timing is generated through the interaction between two cross-listening full-duplex models. At every frame, each instance receives the other speaker's audio and independently decides whether to emit the next script token or remain silent. No global schedule specifies when a turn begins or ends. Instead, the interaction is regulated by three control parameters.

The first parameter is the \emph{handoff window} $W_h$, which defines the interval around the projected end of the current turn during which the listener is allowed to take the floor:

\begin{equation}
\text{handoff allowed}
\iff t\in\left[\tau_{\mathrm{end}}-\frac{W_h}{2},\;\tau_{\mathrm{end}}+\frac{W_h}{2}\right],
\end{equation}

where $\tau_{\mathrm{end}}$ is estimated from the remaining script tokens of the current speaker.

The second parameter, \texttt{max\_padding} ($P_{\max}$), limits the maximum number of consecutive \texttt{PAD} frames before a floor-eligible speaker is forced to begin its queued turn, preventing unrealistically long silent gaps.

The third parameter, \texttt{bc\_refractory} ($R_{bc}$), specifies the minimum interval between two listener backchannels, preventing excessive acknowledgements.

These parameters regulate only the frequency of interaction events. The precise onset of a floor transition, the duration of an overlap, and the attachment point of a backchannel are determined online by the two duplex models in response to
each other's audio. Conversational timing therefore emerges from interaction rather than from an explicit schedule.

\subsection{Fidelity Re-rendering}
\label{sec:m3}

The duplex models generate a symbolic interaction score rather than the final waveform. For each conversational event, the score records the active speaker, turn boundaries, overlap regions, and backchannel positions. This symbolic representation is subsequently rendered into high-fidelity speech using CosyVoice~\cite{cosyvoice}.

Since the duplex model and the TTS system generally speak at different rates, absolute timestamps cannot be copied directly. Instead, timing is transferred using relative positions.

For a silent transition, the silence duration

\[
g=\frac{a_{u+1}-b_u}{f}
\]

is preserved directly so that the rendered turn begins $g$ seconds after the previous rendered turn ends.

For an overlapped transition, the relative entry position is computed as

\begin{equation}
e = \frac{a_{u+1}-a_u}{b_u-a_u},
\end{equation}

where $e$ denotes the fraction of the current turn that has elapsed when the next speaker begins. The rendered overlap is reproduced by placing the onset of turn $u+1$ at $eD_u$, where $D_u$ is the rendered duration of turn $u$. Backchannels are mapped identically according to their relative position within the hosting turn. Consequently, overlap structure is preserved regardless of differences in speaking rate between the duplex model and the renderer.

Three implementation details further improve rendering quality. A sliding prompt maintains speaker consistency throughout long conversations. Interrupted speech is physically truncated with a short crossfade to produce acoustically real interruptions. Backchannels are synthesized independently and mixed into the main conversation at their mapped positions to reproduce concurrent speech.

Because the symbolic interaction score completely specifies the generated conversation, annotations are obtained automatically during synthesis rather than through post-processing. Word timestamps are produced directly from CosyVoice alignment, speaker activity is exported as RTTM segments, and overlap, interruption, gap, and backchannel labels are read directly from the symbolic score.

% ==========================================================================
\section{Related Work}

\subsection{Scripted conversational speech synthesis}

Recent advances in LLMs and neural TTS have enabled large-scale conversational speech synthesis. Existing systems typically generate a dialogue script with an LLM and synthesize each utterance using a neural speech generator. Behavior-SD~\cite{behavior_sd} incorporates behavior prompts to control conversational events, while PersonaPlex~\cite{personaplex} further models speaker personas and dialogue roles. Other dialogue synthesis systems similarly introduce overlap, interruption, or backchannel events through dialogue markers, behavior tags, or handcrafted timing rules~\cite{fdxdata,behavior_reason,fdb_lin,fdb_peng}. These approaches provide fine-grained control over dialogue semantics, making them effective for scalable corpus construction and controllable dialogue generation.

\subsection{Full-duplex conversational modeling}

Full-duplex conversational models directly model bidirectional spoken interaction. dGSLM~\cite{dgslm} first demonstrated dual-stream spoken dialogue generation, while Moshi~\cite{moshi} introduced an inner-monologue architecture that enables real-time duplex conversation. More recent systems, including SyncLLM~\cite{syncllm} and SALMONN-Omni~\cite{salmonn}, further improve synchronous conversational modeling and naturally produce interruptions, and other conversation dynamics through continuous listening while speaking. These models demonstrate that realistic interaction can emerge directly from duplex communication rather than explicit scheduling.

These two research directions address complementary aspects of conversational speech generation: one emphasizes controllable dialogue generation, while the other emphasizes interaction modeling. Our work combines the strengths of both by explicitly separating content generation, interaction generation, and acoustic rendering into three independent stages.

% ==========================================================================
\section{Experiments}
% ==========================================================================
Unless noted, real-dialogue reference statistics are drawn from PriMock57~\cite{primock57} together with published clinical-dialogue statistics; the reference voice for rendering is drawn from the same source but is used only for acoustic consistency, orthogonal to the timing statistics under test. Dialogue scripts are generated with DeepSeek-V4~\cite{deepseek}.

\subsection{Experiment 1: Distributional Realism}
\label{sec:exp1}
We first evaluate whether the proposed duplex interaction produces conversational dynamics that matches real dialogue statistics. Specifically, we compare the generated conversations against two stitching-based baselines using four complementary measures: floor-transfer-offset (FTO) Wasserstein distance, Kolmogorov--Smirnov (KS) statistic, overlapped-transition ratio, and long-tail gap ratio. The reference distribution is computed from $5{,}611$ floor transfers extracted from PriMock57, and is consistent with published turn-taking statistics~\cite{heldner,stivers,levinson}. Emergent statistics in this experiment are computed on general-domain scripts, so that the mechanism is evaluated independently of register effects; the medical-domain instantiation of the same configuration is reported in Table~\ref{tab:exp3} and Table~\ref{tab:exp4tiers}.

\begin{table*}[t]
\caption{FTO distributional realism versus real clinical dialogue (PriMock57,
5,611 transitions). Lower $W_{\mathrm{fto}}$ and KS indicate better agreement with
the real distribution. Best synthetic results are shown in \textbf{bold}.}
\label{tab:exp1}
\centering
\footnotesize
\begin{tabular}{lcccc}
\toprule
Method &
$W_{\mathrm{fto}}\!\downarrow$ &
KS$\!\downarrow$ &
Overlapped transitions (\%) &
Long-tail gaps (\%) \\
\midrule
Real (PriMock57)      & -- & -- & 42.8 & 20.5 \\
Stitching (fixed gap) & 0.783 & 0.557 & 0.0 & 0.0 \\
Stitching (random gap)& 0.695 & 0.432 & 0.0 & 0.0 \\
Ours                  & \textbf{0.366} & \textbf{0.197} & 38.5 & \textbf{30.8} \\
\bottomrule
\end{tabular}
\end{table*}

\begin{figure}[t]
\centering
% NOTE: keep your existing figure file; path below is a stand-in.
\includegraphics[width=\linewidth]{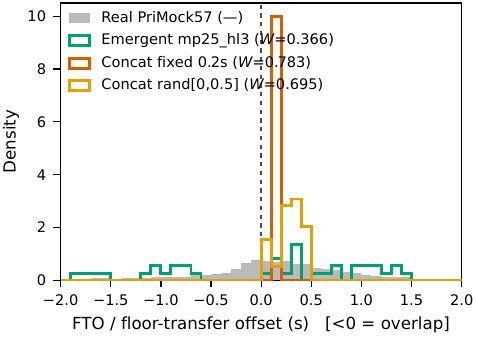}
\caption{Floor-transfer-offset distributions. Emergent output overlaps the real
distribution including its overlapped-transition mode and long silent tail;
stitching collapses to a single near-zero-gap mode.}
\label{fig:fto}
\end{figure}

As summarized in Table~\ref{tab:exp1} and Fig.~\ref{fig:fto}, the proposed method produces conversational dynamics closer to the real-dialogue reference distribution than the stitching baselines evaluated in this study. The FTO Wasserstein distance is reduced from $0.695$ to $0.366$ ($47\%$ relative improvement), while the KS statistic decreases from $0.432$ to $0.197$. More importantly, unlike stitching-based synthesis, the proposed framework naturally reproduces both overlapped transitions and the long-tail silence distribution.

The observed improvement remains stable under multiple robustness checks. Varying the utterance-end estimation constant over $\{0.3,0.4,0.5\}$\,s yields consistent rankings ($W_{\text{fto}}=0.435/0.357/0.292$), which outperform both baselines. Bootstrap evaluation across six random seeds produces a $95\%$ confidence interval of $[0.280,0.545]$, which remains well separated from the baselines. Re-estimating utterance boundaries using an energy-based VAD further improves the agreement with real conversations, reducing $W_{\text{fto}}$ to $0.240$ and
bringing the emergent FTO median to $0.08$\,s, close to the real value of
$0.11$\,s.

\begin{figure}[t]
\centering
% NOTE: keep your existing figure file; path below is a stand-in.
\includegraphics[width=\linewidth]{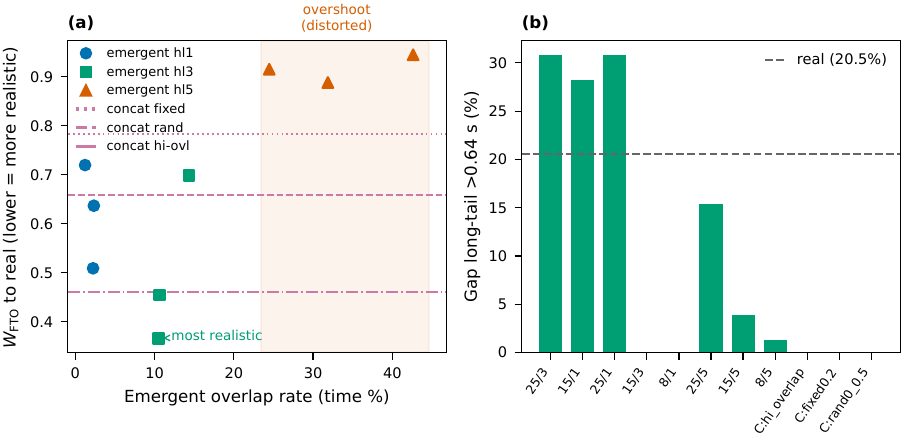}
\caption{Realism as a function of the overlap-control setting. The mechanism's
advantage holds inside the real-dialogue envelope and disappears once the overlap
knob is driven past it---honestly bounding the claim.}
\label{fig:tradeoff}
\end{figure}

The proposed interaction mechanism is most effective within the overlap range observed in natural dialogue. Excessively increasing overlap leads to less realistic conversational timing, indicating that the method is designed to reproduce natural interaction rather than arbitrary overlap patterns.

\subsection{Experiment 2: Fidelity of Decoupled Generation}
The proposed framework explicitly separates content generation, conversational timing, and acoustic rendering. This experiment verifies that information does not unintentionally leak across stages and each stage preserves the decision established by the previous one. Specifically, we evaluate content fidelity, timing preservation, and speaker consistency after the complete re-rendering pipeline.

\begin{table}[t]
\caption{Pipeline fidelity. Content lock, timing transfer, and re-rendering each preserve their invariant.}
\label{tab:exp2}
\centering
\footnotesize
\begin{tabular}{lc}
\toprule
Check & Result \\
\midrule
Forced WER (content lock)          & $0\%$ \\
Overlap-event mapping deviation    & $0$ (exact) \\
Re-rendering WER                   & $\leq 1.7\%$ \\
Speaker consistency (CAM++)        & $0.82$ \\
Truncation validity (per segment)  & $0.01$\,pp \\
Overlap audit, dominant tier       & $0.5$\,pp \\
\bottomrule
\end{tabular}
\end{table}

Table~\ref{tab:exp2} summarizes the fidelity of each stage in the proposed pipeline. Since lexical content is fixed before duplex interaction, the generated transcripts remain identical to the input script, resulting in $0\%$ forced WER. Likewise, overlap events are transferred exactly from the symbolic interaction score to the rendered speech without deviation. Acoustic re-rendering introduces at most $1.7\%$ WER while preserving speaker identity, achieving a CAM++~\cite{campp} similarity of $0.82$. 

The last two rows of Table~\ref{tab:exp2} confirm acoustic-level timing consistency: delivered overlap closely matches the annotations, with the remaining discrepancies explained by low-level ($-6$,dB) backchannel overlays that are annotated separately from floor overlap.

\subsection{Experiment 3: Emergence and Controllability}
To examine whether the proposed interaction remains both adaptive and controllable, we compare conversations generated under different dialogue styles and timing configurations. Since dialogue content is fixed throughout generation, differences in conversational behavior are solely determined by the duplex interaction process. Under identical scripts and knob settings, different random seeds yield distinct performances (zero to three barge-ins per dialogue), confirming that overlap placement is stochastic and interaction-driven rather than deterministic.

\begin{table}[t]
\caption{Register-adaptive overlap under identical knob settings, and backchannel
placement.}
\label{tab:exp3}
\centering
\footnotesize
\begin{tabular}{lc}
\toprule
Condition & Value \\
\midrule
Overlap \% --- medical script     & $7.2$ \\
Overlap \% --- casual script      & $10.5$ \\
Overlap \% --- real clinical      & $6.3$ \\
\midrule
Backchannel at prosodic boundary  & $95.5\%$ \\
\quad baseline                    & $71.7\%$ \\
\bottomrule
\end{tabular}
\end{table}

Table~\ref{tab:exp3} shows that the generated interaction adapts naturally to different dialogue styles while remaining controllable. Under identical timing parameters, medical conversations consistently exhibit less overlap than casual conversations ($7.2\%$ vs.\ $10.5\%$), closely matching the restrained interaction style observed in real clinical dialogue ($6.3\%$). Meanwhile, backchannels emerge at prosodically appropriate positions in $95.5\%$ of cases, substantially exceeding the random baseline ($71.7\%$). Finally, increasing the handoff window consistently increases overlap frequency, whereas increasing \texttt{max\_padding} produces more long silent gaps, allowing conversational dynamics to be adjusted smoothly from polite to adversarial interaction.

\subsection{Experiment 4: Downstream Stress Test}
\label{sec:exp4}
We evaluate whether the generated conversations provide a controllable benchmark for overlapping speech recognition. The released benchmark contains $180$ segments ($175$ minutes) spanning three interaction difficulty levels, constructed by varying the timing parameters described in Section~\ref{sec:m2}. As summarized in Table~\ref{tab:exp4tiers}, the three tiers exhibit progressively denser interaction patterns, with the natural tier closely matching the overlap  statistics of real clinical dialogue.

\begin{table}[t]
\caption{The three tiers separate monotonically on interaction density; the natural tier closely matches the time-overlap rate of real clinical dialogue. Time overlap and long-tail from construction-time ground truth.}
\label{tab:exp4tiers}
\centering
\footnotesize
\begin{tabular}{lcccc}
\toprule
Tier & Overlap \% & Overlap.\ trans.\ \% & Long-tail \% & BC/min \\
\midrule
Polite       & $2.0$  & $11.6$ & $34.4$ & $3.9$ \\
Natural      & $6.6$  & $20.6$ & $28.8$ & $3.7$ \\
Adversarial  & $12.2$ & $27.2$ & $21.0$ & $2.6$ \\
Real         & $6.3$  & $42.8$ & $20.5$ & $1.5$ \\
\bottomrule
\end{tabular}
\end{table}

We evaluate Whisper large-v2~\cite{whisper} and wav2vec2~\cite{wav2vec2} on the three difficulty tiers, reporting overlap-region WER separately from the overall WER. As shown in Table~\ref{tab:exp4b2}, overlap-region errors are consistently $3\text{--}7\times$ higher than aggregate WER and increase monotonically with interaction difficulty for both systems, whereas clean-region performance remains nearly unchanged. These results indicate that the proposed timing controls effectively regulate ASR difficulty through conversational overlap rather than overall speech quality.

\begin{table}[t]
\caption{ASR WER (\%) on corrected audio. Overlap-region error far exceeds aggregate and rises monotonically with difficulty across both systems, while clean regions stay flat---degradation is overlap-driven and controllable.}
\label{tab:exp4b2}
\centering
\footnotesize
\begin{tabular}{llccc}
\toprule
System & Region & Polite & Natural & Advers. \\
\midrule
\multirow{3}{*}{Whisper}  & aggregate      & $4.1$  & $6.8$  & $12.7$ \\
                          & overlap-region & $28.3$ & $48.1$ & $56.7$ \\
                          & clean          & $3.1$  & $1.8$  & $2.2$  \\
\midrule
\multirow{3}{*}{wav2vec2} & aggregate      & $16.5$ & $20.2$ & $24.5$ \\
                          & overlap-region & $50.7$ & $64.8$ & $71.9$ \\
                          & clean          & $15.2$ & $14.7$ & $13.3$ \\
\bottomrule
\end{tabular}
\end{table}

To distinguish the effects of conversational timing and acoustic rendering, we further conduct a controlled $2\times2$ ablation with matched overlap rates (Table~\ref{tab:exp4b3}). Note that the absolute WER values in Table~\ref{tab:exp4b3} are not directly comparable to the by-tier numbers in Table~\ref{tab:exp4b2}: the ablation cells hold the overlap rate and renderer fixed and exclude backchannel overlays, so their overlap regions consist only of turn-transition overlaps, a different and acoustically simpler composition than the full benchmark tiers. The comparison of interest is therefore \emph{within} the $2\times2$, not across tables. Holding rendering fixed, emergent and stitched timing produce comparable overlap-region WER, suggesting that conversational timing itself has only a limited influence on downstream ASR performance. In contrast, replacing truncate-and-fade rendering with equal-volume overlap substantially increases recognition difficulty, indicating that rendering intensity dominates overlap-region ASR performance.

\begin{table}[t]
\caption{$2\times2$ ablation of overlap-region WER (\%, $95\%$ CI). Rendering, not
timing, drives downstream difficulty; all synthetic cells fall well below real.}
\label{tab:exp4b3}
\centering
\footnotesize
\begin{tabular}{llc}
\toprule
Timing & Rendering & Overlap-region WER \\
\midrule
Emergent  & truncate$+$fade      & $18.8$ \;($13.1$--$24.6$) \\
Stitched  & truncate$+$fade      & $23.2$ \;($15.5$--$31.8$) \\
Emergent  & equal-volume         & $44.1$ \;($37.5$--$49.6$) \\
Stitched  & equal-volume         & $43.3$ \;($36.9$--$50.0$) \\
\midrule
Real      & ---                  & $76.5$ \;($56.4$--$99.6$) \\
\bottomrule
\end{tabular}
\end{table}

Two effects separate cleanly. \textbf{Timing has no detectable effect}: with rendering held fixed, emergent and stitched confidence intervals overlap almost entirely ($18.8$ vs.\ $23.2$; $44.1$ vs.\ $43.3$), so at $n\!=\!30$ any effect of emergent versus authored timing on ASR error is small relative to rendering. \textbf{Rendering dominates}: with timing held fixed, replacing truncate-and-fade with equal-volume overlay roughly doubles the error (disjoint CIs). Counter-intuitively, the acoustically \emph{cleaner} truncate-and-fade rendering makes overlap \emph{easier}: fading the interrupted speaker lets the recognizer lock onto the surviving barge-in, moving error away from the real value. The harder, equal-volume condition is closer to real but still leaves a large gap---every synthetic cell sits well below the real overlap-region WER of $76.5\%$.

\textbf{Robustness.} The observed trend generalizes across different speaker pairs and dialogue domains. Re-rendering with $12$ LibriTTS~\cite{libritts} speaker pairs preserves the overlap-region degradation, and similar behavior is observed on out-of-domain casual dialogue. In contrast, speaker diarization error (pyannote-3.1~\cite{pyannote}) remains largely unchanged across difficulty levels, indicating that the benchmark primarily stresses speech recognition rather than speaker attribution.

% ==========================================================================
\section{Limitations}
\label{sec:limits}
% ==========================================================================
The proposed framework has several limitations. First, conversational interaction is only partially emergent. Although turn timing is generated online through duplex interaction, global interaction statistics such as overlap frequency and backchannel density are still regulated by explicit control parameters. Likewise, overlap is currently confined to turn-transition regions and does not model deep mid-utterance interruptions.

Second, the strict decoupling between content and timing inevitably limits conversational adaptability. Since dialogue content is fixed before duplex generation, speakers cannot revise or abandon planned utterances after being interrupted, preventing richer interaction phenomena such as repair, self-correction, or content negotiation.

Finally, the current benchmark remains a simplified approximation of real clinical conversations. It is constructed from a single reference corpus and synthetic rendering, and downstream ASR difficulty is influenced more by acoustic rendering than by emergent timing itself. Although the proposed benchmark provides controllable conversational dynamics, further improvements in overlap rendering and multi-speaker interaction are needed to better match real-world conversations.

% ==========================================================================
\section{Conclusion}
% ==========================================================================
We presented Agentic-DuplexGen, a zero-training pipeline that decouples content, timing, and acoustics so that overlapping conversational speech can be scripted for content yet emergent in timing and high-fidelity in sound. Emergent turn-taking is distributionally closer to real clinical dialogue than stitching and reproduces overlapped transitions and long gaps that are absent from the evaluated stitching baselines, while three physical knobs give controllable densities with a clean control/emergence boundary. The resulting three-difficulty medical benchmark ships with construction-time ground truth for stress-testing ASR and diarization.

\section*{Ethics and Data Statement}
The reference corpus (PriMock57) is used under its license. All released audio is synthetic; we label it as such and restrict its use to research on speech recognition and dialogue robustness. The synthetic data carries no clinical validity claim and must not be used for medical decision-making.

\section*{Acknowledgment and AI Disclosure}
This anonymous draft follows the IEEE submission requirements. Claude and Grammarly were used to polish drafts and assist with table organization in this manuscript. The experimental design and results, claims, citations were run and written by the authors.

\bibliographystyle{IEEEtran}
\bibliography{ref}

\end{document}